\documentclass[english]{paper}
\usepackage[T1]{fontenc}
\usepackage[latin9]{inputenc}
\usepackage[letterpaper]{geometry}
\usepackage{amsmath}
\usepackage{graphicx}
\usepackage{amssymb}

\makeatletter

\providecommand{\tabularnewline}{\\}

\makeatother

\usepackage{babel}

\begin{document}

\title{Travel Time Estimation Using Floating Car Data}

\author{Raffi Sevlian, Ram Rajagopal}

\maketitle
This project explores the use of machine learning techniques to accurately
predict travel times in city streets and highways using floating car
data (location information of user vehicles on a road network). The
aim of this report is twofold, first we present a general architecture
of solving this problem, then present and evaluate few techniques
on real floating car data gathered over a month on a 5 Km highway
in New Delhi.

\section{Floating Car Data Based Traffic Estimation}

Data used to estimate the travel times on road networks come in two
varieties, one being fixed sensors on the side of the road such as
magnetometer detectors or highway cameras {[}5, 6{]}. The second method
is floating car data (FCD). Floating car data are position fixes of
vehicles traversing city streets throughout the day. The most common
type of FCD comes from taxi's or delivery vehicles which are on main
arterial roads and highways throughout most of the day. 

This second approach has many positive and negative attributes that
must be dealth with to provide accurate travel time inference. First,
FCD is perhaps the most inexpensive data to attain, since many taxi
services, and delivery companies automatically gather this data on
their vehicles for logistic purposes. Second, position fixes are generally
very accurate, since GPS is used and this has high accuracy. There
are however many disadvantages as well. First, FCD is usually sampled
infrequenty, on the order of 2-3 minutes. The reason for this is that
taxi or delivery companies do not need such fine time granularity
of their vehicles position. Therefore quiet a bit of preprocessing
needs to take place in order to {}``snap'' sets of points onto the
proper streets with the possibility that multiple paths might have
lead to the same pair of traverse points. Another disadvantage of
this method is that a high density of data is required to get meaningfull
travel time predictions for a given road network. Keeping all this
in mind, constructing more and more accurate travel time predictions
can be a fruitful Algorithms/Machine Learning/Statistical Modelling
problem with various problems to tackle.

\section{Prediction Architecture}

\begin{figure}
\label{fig:Architecture}\includegraphics[scale=0.5]{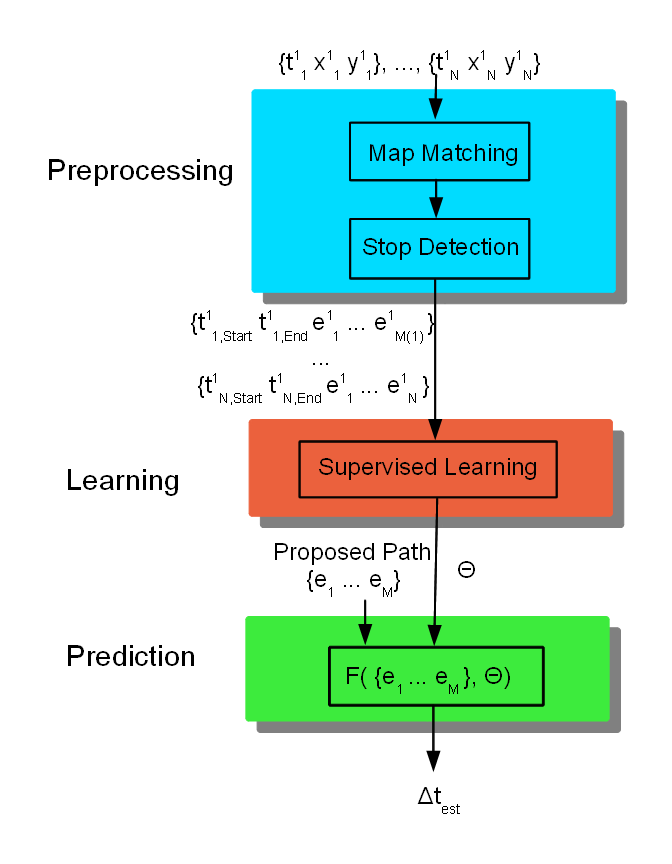}\caption{Travel Time Prediction Architecture: For single user, input is sequence
of $\left\{ t_{1}x_{1}y_{1}\right\} ...\left\{ t_{N}x_{N}y_{N}\right\} $. }

\end{figure}

Building a traffic estimation, system using millions of incoming FCD
streams is a computational and algorithmic challenge. Various subcomponents
requiring significant research work need to be developed and integrated.
Here we give a breif overview of the various processing steps required
as well as the issues explored in this report. Figure (\ref{fig:Architecture})
presents a high level view of such a system.

\section{Preprocessing}

\subsection{Motion Detection}

Since the incoming streams of traffic data come from taxi cabs, delivery
vehicles, or other comercial fleet vehicles, there will always be
a certain amount of ambiguity between a slowdown in traffic and a
commercial stop by the vehicle. (i.e. for a taxi customer, or a delivery
vehicle dropping off packages). Therefore, any further processing
must clean out all unwanted stops that are in the GPS logs. The most
common and inuitive technique, and that which is used in this project
is to track the the number of consequtive GPS points that are less
than a specified distance $D_{max}$ from each other.

At each iteration a running tally $N_{t}$ is kept for the number
of GPS points that are within $D_{max}$ of each other. If $N_{t}$
is greater than some threshold $N_{max}$ then the previous $N_{max}$
points are labeled as invalid. To compute the threshold parameters
$\left(N_{max}D\right)$, a supervised learning algorithm is can technique.
In live taxi data, the taxi driver will activate a counter to track
the distance crossed by the vehicle and the time spent farying the
customer accross town, this is used to construct the proper labels
for motion and stopping of the taxi. If this secondary informaiton
is not available, another option is to visually inpect a large data
set of GPS data, and manually identify the regions where a stop has
taken place.

\subsection{Map Matching}

Map Matching is a widely studied problem in transportation research
is perhaps the most computaionally difficult and important subcomponent.
The input is a time indexed sequence of GPS coordinates $\left\{ t_{1}x_{1}y_{1}\right\} ...\left\{ t_{N}x_{N}y_{N}\right\} $,
where $t_{i}$ is a standard unix timestamp data structure, and pairs
$\left(x_{i}y_{i}\right)$ represent latitude and longitude coordinate.
Map matching attempts to efficiently and accurately match consecutive
points to sequence of links representing road segments in a stardard
city map. Standard map matching outputs, $\left\{ t_{1,Start}\, t_{1,End}\,\left\{ e_{n}^{1}\right\} _{n\in E_{1}}\right\} $
$\ldots$ $\left\{ t_{N,Start}\, t_{N,End}\left\{ e_{n}^{N}\right\} _{n\in E_{1}}\right\} $,
here$\left\{ e_{n}^{i}\right\} _{n\in E_{1}}$ represents a set road
segment links for the $i^{th}$ path inthegral. Also, $E_{i}$ is
the set of all indices representing such path edges. Figure (\ref{fig:RoadNetwork})
shows a simple example of map matching.

Map matching is done in a variety of ways depending on a tradeoff
of accuracy and efficiency. For the task of realtime high capacity
map matching, several techniques involving hueristic graph search
are shown in {[}1, 2, 3{]}. These techniques follow a general strategy
of greedily adding road segments to a solution set as points are processed,
each candidate road segment receives a score by a distance function
defined on road segments and GPS probes. A standard distance function
is is given by finding the shortest distance between the GPS point
and point on the road segment. 

Given a road segment defined by all convex combinations of two GPS
coordinates, $A$ and $B$. $e_{A,B}=\left\{ \left(x,y\right):\alpha\in\left[0,1\right],\left(x,y\right)=\alpha A+\left(1-\alpha\right)B\right\} $
points inside the road segment are defined as, $e_{A,B}\left(\alpha\right)=\alpha A+\left(1-\alpha\right)B$.
With the standard distance function used map matching algorithms is. 

$\qquad$

$d\left(\left\{ xy\right\} ,\: e\right)=\left\{ {\underset{\alpha}{\min}d_{GPS}\left(e_{A,B}\left(\alpha\right),\left\{ xy\right\} \right)\qquad\measuredangle\left(\left\{ xy\right\} ,A\right)<\frac{\pi}{2}\; and\;\measuredangle\left(\left\{ xy\right\} ,B\right)<\frac{\pi}{2}\atop \min\left\{ d_{GPS}\left(A,\left\{ xy\right\} \right),d_{GPS}\left(B,\left\{ xy\right\} \right)\right\} \qquad else}\right.$

$\qquad$

Where $d_{GPS}\left(p_{1\,}p_{2}\right)$and $\measuredangle\left(p_{1},\, p_{2}\right)$
represent standard geographic distance and angle between two GPS coordinates.
After the candidate pool grows to a specified size, pruning via a
hueristic shrinks the candidate pool.

\section{Supervised learning for Traffic Estimation}

This report explores the use of supervised learning for traffic inference
on links on a road network. Each technique relies on different modelling
assumptions and produces different estimation results and has particular
advantages and disadvantages that dictates use in individual situations. 

The most general form of learning and prediction is that after learning
takes place, some model parameterized by $\Theta$ will be available.
Some prediction function, $F\left(\left\{ e_{n}^{i}\right\} _{n\in E_{i}};\:\Theta\right)$
will use the learned parameters, and output an estimate of the travel
times through a new set of links $\left\{ e_{n}^{i}\right\} _{n\in E_{i}}$.

\subsection{Regression Technique:}

\begin{figure}
\label{fig:RoadNetwork}\includegraphics[scale=0.55]{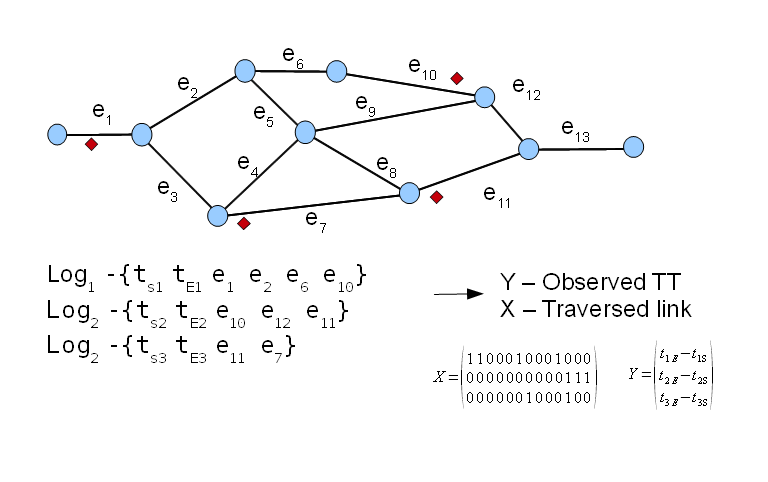}\caption{Example of map matching of set of GPS points. Input is set of points
with outputted path integrals.}

\end{figure}

Regression techniques rely on an additive model for travel times.
That is the travel time for a commuter traversing a set of links will
be the sum of the travel times for the set of links traversed. Therefore
if the current link travel times are already known, then the estimated
travel time for some path through the network in figure (\ref{fig:RoadNetwork})
will the sum of the link travel times. 

$F\left(\left\{ e_{n}^{i}\right\} _{n\in E_{i}};\:\Theta\right)=\sum_{n\in E_{i}}t_{\Theta}\left(e_{n}\right)$

Under a regression model, each $t_{\Theta}\left(e_{i}\right)=\theta_{i}$
therefore given link travel times, any observed travel time will follow 

$y^{i}|\sum_{n\in E_{i}}t_{\Theta}\left(e_{n}\right)\sim N\left(\sum_{n\in E_{i}}t_{\Theta}\left(e_{n}\right),\:\sigma^{2}\right)$

$Y=X\Theta+E$

Throughout this project, we make different assumptions on the observed
travel times depending on whether long term averages, or short term
deviations are being computed.

\subsubsection{Long Run Historical Travel Time}

A goal of traffic flow analysis is having long term historic data
on flow volume through various links. To construct a regression model
taking into account spatial variation of historic TT values, we can
assume that adjacent links are distributed normally: 

$t\left(e_{n}\right)-t\left(e_{n-1}\right)\sim N\left(0,\:\tau^{2}\right)\; n=\left\{ 1,\ldots,N-1\right\} $

This enforces a penalty on sudden changes to the historic means, as
well as makes estimation of parameters tractable when the maximum
likelihood estimator is underdetermined. With this assumption we construct
the following maximum a posteriori estimator.

$\Delta=\underset{t\left(e\right)}{\arg\max}\sum_{i}log\: P\left(y^{i}|\sum_{n\in E_{i}}t\left(e_{n}\right)\right)+\sum_{m}log\: P\left(t\left(e_{m}\right)-t\left(e_{m-1}\right)\right)$

$\quad=\underset{t\left(e\right)}{\arg\min}\sum_{i}\frac{1}{\sigma^{2}}\left\Vert y^{i}-\sum_{n\in E_{i}}t\left(e_{n}\right)\right\Vert _{2}^{2}+\frac{1}{\tau^{2}}\left\Vert \sum_{m}t\left(e_{m}\right)-t\left(e_{m-1}\right)\right\Vert _{2}$

$\quad=\underset{\Theta}{\arg\min}\left\Vert Y-X\Theta\right\Vert +\lambda_{1}\left\Vert D\Theta\right\Vert _{2}$

Which is simply a Ridge Regression or L2 Regression with smoothness
penalty on the variation of historic mean along different links {[}6{]}.

\subsubsection{Short Term Incidence Detection on Live Data}

Given a short time window (on order of 1 hour) we assume the distribution
on any given day for link $t\left(e_{i}\right)=\theta_{i}+\Delta_{i}$where
$\theta_{i}$is a historical average and $\Delta_{i}$is random variable
representing the deviation that day from the historical average. The
important assumption here is that this deviation is distributed with
a heavy tail. (i.e. no deviation with high probability and large deviation
with low probability). To model this, we use a laplacian prior distrbution:
$p(\Delta_{i};\sigma)=e^{|\frac{-\Delta}{\sigma}|}$. 

Given a set of historic data, and a small number of measured daily
measured paths, we can use this model to computer short term deviations
from the historic values.

$\hat{\Delta}=\underset{t_{\Theta}\left(e\right)}{\arg\max}\sum_{i}log\: P\left(y^{i}|\sum_{n\in E_{i}}t_{\Theta}\left(e_{n}\right)\right)+\sum_{m}log\: P\left(t_{\Theta}\left(e_{m}\right)\right)$

$\quad=\underset{t_{\Theta}\left(e\right)}{\arg\min}\sum_{i}\frac{1}{\sigma^{2}}\left\Vert y^{i}-\sum_{n\in E_{i}}t_{\Theta}\left(e_{n}\right)\right\Vert _{2}^{2}+\frac{1}{\tau^{2}}\sum_{m}\left\Vert t_{\Theta}\left(e_{m}\right)\right\Vert _{2}^{2}$

$\quad=\underset{\Delta}{\arg\min}\left\Vert Y-X\left(\Theta+\Delta\right)\right\Vert _{2}^{2}+\lambda_{2}\left\Vert \Theta+\Delta\right\Vert _{1}$

Therefore in every small time window that live probe data arrives,
an L1 regression is performed estimate deviations from the historic
travel times. Since probe volume is small compared to the number of
links, this technique makes intuitive sense. When performing live
prediction we assume that the parameters $\left\{ \Theta,\,\lambda_{1},\,\lambda_{2}\right\} $
are already computed. Therefore in each time window used for regression,
the parameters $\left\{ \Delta\right\} _{i}$ are computed and predictions
are given by $\sum_{n\in E_{i}}\left(\theta_{n}+\Delta_{n}\right)$.

\subsubsection{Computing Model Parameters}

\begin{figure}
\label{fig:DataModel}\includegraphics[scale=0.4]{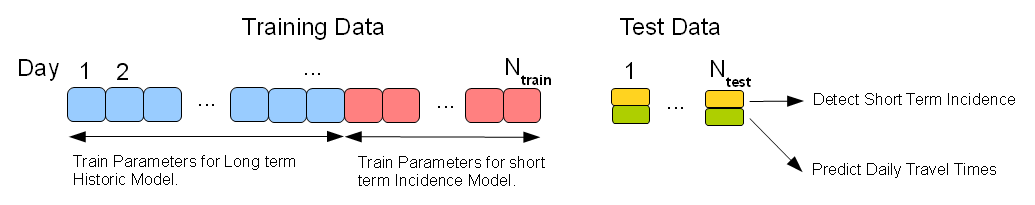}

\caption{Data Processing model for estimating historic means for link travel
times, as well as L1 regression parameters. Test data used daily to
compute short term deviations from historic average.}

\end{figure}

In modelling travel time distributions, we explicitley separate long
term and short term behaviour to easily estimate the model parameters.
With this, parameter estimation is performed in two stages. First,
a subset of the training data is aggregated and crossvalidation is
performed to determine $\Theta$ and appropriate $\lambda_{1}$. In
the second stage, the short term model parameter $\lambda_{2}$ is
computed with similar cross validation with training data (See note
(\ref{fn:EM_note})). This data processing model is shown in figure
(\ref{fig:DataModel}). This is a visual interpretation of how data
processing takes place. Each block is a tuple of paths and travel
times (i.e. X, and Y from figure (\ref{fig:RoadNetwork})). 

\footnote{\label{fn:EM_note}Recall that this separation between long term historic
data and short term fluctuations is an assumption we explicitly make.
A more thorough model would merely include daily laplacian deviations
into the first model. However comstructing an estimator for this latent
variable model, requires an EM algorithm. This is a definately a future
extension of our work.%
}

\subsection{Travel Time Median Backprojection:}

The second general technique used in travel time estimation is to
merely backproject the travel time distances across each traversed
link weighted proportionally to the lenght of the links. After many
paths are processed, each link will map to a list of backprojected
travel time values. The travel time estimate of the link is simply
the median over the distribution. We use this algorithm as a naive
baseline to compare the more sophisticated model.

\section{Experimental Results}

\subsection{Preprocessing Results}

\begin{figure}
\label{fig:RoadExperiment}\includegraphics[scale=0.4]{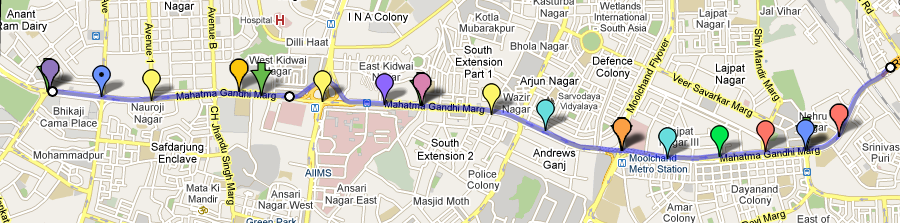}\caption{Single road learning experiment.}

\end{figure}

Components of the system architecture were implemented according to
figure (\ref{fig:Architecture}) to process floating car data from
approximately 40 Taxi's over the course of about one month in urban
New Delhi. The total dataset of GPS logs span 22 days: dates 4/5/2008
to 4/26/2008. The data was sampled infrequently at about 2-3 minutes.
The reason for this is that some GPS devices sampled every 2 minutes
while some sampled every 3 minutes. Since the focus of this report
is the core inference techniques, a single 5 Km highway (figure (\ref{fig:RoadExperiment}))
was chosen for analysis instead of a region of the city, or the entire
city. instead of an entire city. Another reason why a long highway
was chosen is that the fundemental assumption that link travel times
can be added to estimate the combined travel time is a more accurate
model for highways than for city arterial roads {[}4{]}.

\begin{table}
\label{tag:Table_Data_Stats}\caption{Input Data Statistics: 4/5/2008 - 4/26/2008 (22 days, 8-9 AM)}
\begin{tabular}{|c|c|c|c|c|}
\hline 
min & max & mean & std & \tabularnewline
\hline
\hline 
34 & 110 & 66.3 & 27.5 & Raw Data in Sector\tabularnewline
\hline 
11 & 65 & 26.8 & 13.6 & Processed Path Integrals\tabularnewline
\hline
\end{tabular}
\end{table}

The data included all FCD logs that were contained in the geospatial
coordinates shown in  figure (\ref{fig:RoadExperiment}). Therefore
much of the data belonged to taxi's that might have traversed nearby
streets. To isolate the data required for inference of the highlighted
street, full map matching on the data was implement and used to seperate
FCD data traversing nearby streets. Additionally, stop detection as
described earlier was used to further remove artifacts that would
cause false positives in incident detection. Statistics on the raw
input data and final post processed data are given in table (\ref{tag:Table_Data_Stats}).
For this project, additional processing was done to remove any artifacts
caused by missed detections of taxi stops. From table (\ref{tag:Table_Data_Stats})
it is apparent that a significant proportion of the data contained
stops or belonged to paths on nearby streets. 

\begin{figure}
\label{fig:PostProcess_Example}\includegraphics[scale=0.2]{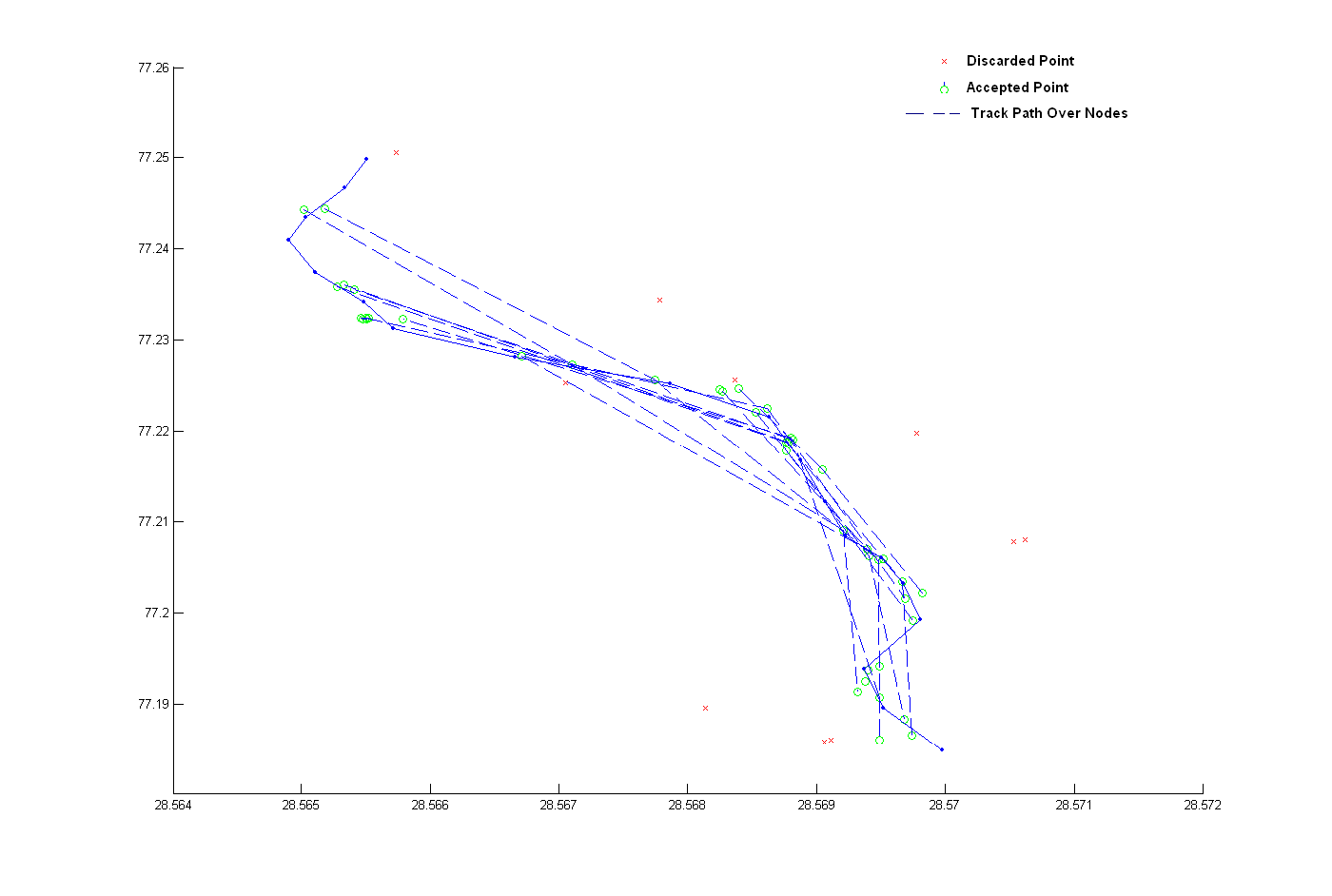}\caption{Track Creation process - Primary reprocessing step.}

\end{figure}

Figure (\ref{fig:PostProcess_Example}) shows a sample of the processing
done. The each green point was determined an acceptable path integral
end point, while red points were rejected. The dashed line indicates
the start and end of each path integral. All logs for this example
were sampled uniformly. By visual inspection, one can infere that
the upper left quadrant will have much lower travel times, than the
lower right portion of the street. 

\subsection{Inference Results}

\begin{figure}
\label{fig:param_fitting}\includegraphics[scale=0.4]{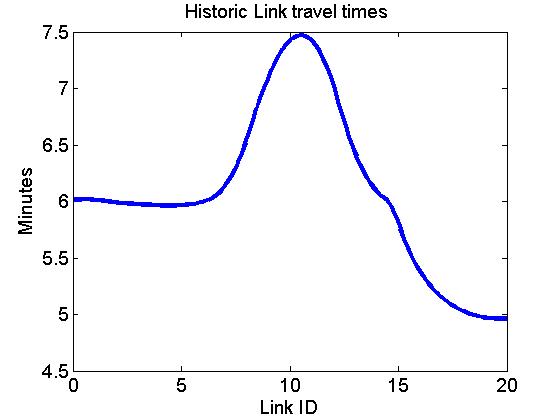}\includegraphics[scale=0.5]{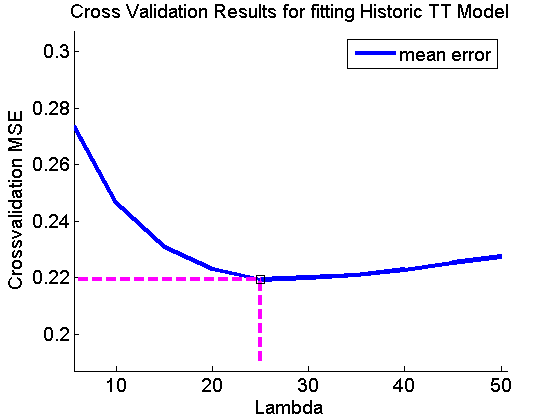}
\end{figure}

Supervised learning of post processed data follows the model presented
in figure (\ref{fig:DataModel}). Here, $N_{train}^{1}$ blocks (each
representing one day's paths) were chosen to learn the historical
travel time model. From the entire $N_{train}^{1}$ 5 fold cross validation
was used with varying model parameter $\lambda_{1}$ to estimate the
optimal historical travel time $\Theta.$ That is the $\lambda_{1}$
yielding the lowest cross validation error for the data was chosen
as the optimal parameter. Figure (\ref{fig:param_fitting}) shows
the results from cross validation; the historical travel times, optimal
$\lambda_{1}$and minimum training error. Cross validation error for
training historic travel times yielded an optimal $\lambda_{1}=24.45$
and $MSE_{min}=22\%$. 

Fitting the parameters for the daily incidence model follows an identical
methodology as for the long term historic mean estimation. $N_{train}^{2}$
days of data are used, wher for each day, leave one out cross validation
(since the number of path integrals is on the order of 10 - 15) is
performed where parameter $\Delta$ is estimated with the training
data and used to predict the travel time of the one out path integral/TT.
As before, $\lambda_{2}$ is varied and an optimal value can be determined
by looking at the lowest test error.

\subsection{Test Prediction}

Finally with the model parameters computed, the incident detection
technique is tested as follows: All data not used in the first two
phases ($N_{test}$) is tested for each individual day, whereby all
but one of the path integrals is assumed to be the day's probe data
(i.e. leave one out model validation) . This data is used to to find
the sparse deviations from the historic mean $\left\{ \Delta\right\} $
by solving the LASSO problem presented earlier. The deviations are
then incorporated with the historic mean to provide the predicted
travel time.

\begin{table}
\label{tab:Table_Final_Results}\caption{Main Results comparing 3 algorithms.}
\begin{tabular}{|c|c|c|c|c|}
\hline 
Technique & Error Rate (\% of True) & Std Dev. & N & $95\:\%$ Confidence Interval\tabularnewline
\hline
\hline 
Historical Average & 26.82 \% & .10 & 234 & $\left[25.53,\:28.1\right]$\tabularnewline
\hline 
Historical Average And Incidence Detection & 22.1 \% & .12 & 234 & $\left[20.56,\:23.6\right]$\tabularnewline
\hline 
Median Backproject & 32.4 \% & .14 & 234 & $\left[30.60,\:34.19\right]$\tabularnewline
\hline
\end{tabular}
\end{table}

In this report we tested 3 separate algorithms to compare their results.
Multiple random assignments were made between the 22 files and the
subsets used for training and testing ($N_{train}^{1},\: N_{train}^{2},\: N_{test}$).
First we merely use the historic means as a default estimate, without
updating the link travel times by any live data. Second we apply the
incidence detection technique introduced in the paper. Finally we
apply the naive median backproject technique. Table (\ref{tab:Table_Final_Results})
summarizes the final computed test errors for the different techniques
discussed. The results show an improvement of incidence detection
over merely using the historical average via regularized least square.
The worst technique was the median backproject. The error values are
in percentage of true link travel time. A simple statistical analysis
of the incidence detection results shows that the improvement is statistically
significant. Under $95\%$ confidence interval there is a statistically
significant difference in each algorithms results.

\section{Conclusions and Future Directions}

This report presents a unified architecture for Floating Car Data
based travel time inference as well as proposes an incidence based
inference technique that is evaluated and shown to be quiet effective
at estimating travel times on various links in transportation networks.
The largest impediment in achieving more accuracy appears to be the
amount of data required. The road network was chosen since from the
data available, it had the highest density of FCD. Regardless, at
the rush hour time of 8-9 AM an average of 26 or so path integrals
were constructed. Therefore future tests require a higher density
of FCD. Also, a unified EM algorithm (mentioned in the footnotes)
for training the historical data, while incorporating the sparse deviations
model should be investigated.

\end{document}